\documentclass[a4paper,11pt,twocolumn,twoside]{article}
\usepackage[dvips]{graphicx}
\usepackage{sepln}

\usepackage{fullname_esp}
\usepackage[utf8]{inputenc}
\usepackage{xcolor}
\usepackage{lmodern}
\usepackage{algorithm} 
\usepackage{algpseudocode} 
\usepackage{url}
\usepackage{amsmath}

\input epsf

\title{Automatic detection of diseases in Spanish clinical notes combining medical language models and ontologies}

\author {\textbf{León-Paul Schaub Torre$^1$}, \textbf{Pelayo Quirós$^1$}, \textbf{Helena García-Mieres$^1$}\\
$^1$ CTIC Centro Tecnológico\\
\{leon.schaub;pelayo.quiros\}@fundacionctic.org, helenagmieres@gmail.com\\
}

\seplnkey{multiclass classification, language model, biomedical, ontology, hybrid method}

\seplnabstract{In this paper we present a hybrid method for the automatic detection of dermatological pathologies in medical reports. We use a large language model combined with medical ontologies to predict, given a first appointment or follow-up medical report, the pathology a person may suffer from. The results show that teaching the model to learn the type, severity and location on the body of a dermatological pathology, as well as in which order it has to learn these three features, significantly increases its accuracy. The article presents the demonstration of state-of-the-art results for classification of medical texts with a precision of 0.84, micro and macro F1-score of 0.82 and 0.75, and makes both the method and the data set used available to the community.}

\firstpageno{1}

\begin{document}



\label{firstpage} \maketitle

%

\section{Introduction}

The digitization of medical records (EHR for electronic health records) is an international initiative that has been in development for decades. One of the first protocols for digitization of reports is ISO TC 215 \footnote{https://www.iso.org/committee/54960.html} created in 1998. It is a norm whose objective is to standardize the digitization of reports from more than 50 countries, including Spain, both visual (radiology, ultrasound, etc.) and textual. This is why it is possible to have a context and a history of each patient, as well as to ease follow-ups. However, the acceleration of this digitization in the last 15 years, with the globalization of high-speed Internet and massive storage capacities, has led to a growth in the amount of available data. For that reason, Natural Language Processing (NLP) has great potential as a tool to help physicians monitoring patients, pre-analyzing EHRs, recognizing entities (NERs), or predicting the pathologies that a person suffers from. In parallel, advances in deep learning \cite{9075398} since the 2010s and the transformers as of 2018 \cite{NIPS2017_3f5ee243} have enabled the creation of more accurate models \cite{schaub2023inteligencia}. Combining both advances, in recent years pre-trained language models specialized in medical vocabulary have been developed and tuned (\textit{fine-tuned}) for the mentioned applications \cite{DeFreitas2021,9903583,9964038}. In Spanish language and in any language other than English \cite{Neveol2018} existing resources are limited, but models such as those developed by \cite{PLN6403,aracena-etal-2023-pre} achieve results comparable to English language models for information extraction tasks \cite{rojas-etal-2022-clinical}, as NER.
\newline
Nevertheless, few papers focus on the prediction of a disease within a clinical report \cite{app122211709}. There are word embedding works that have been successful in connecting a report and a concept (e.g., disease or type of disease) \cite{Araki2023,DBLPjournals/corr/abs-2107-03134}, managing to outperform ontology and semantics works in the last decade \cite{10.1136/amiajnl-2012-001376,BUCHAN201723}. Despite this, there are hardly any reference corpora to have both medical reports in Spanish and the associated pathology, having identified as the only reference the CARES corpus \cite{CHIZHIKOVA2023106581}, although it is focused on radiological data. Neither has a state-of-the-art method been detected that is capable of predicting to which pathology(ies) a given medical report corresponds.


On the other hand, the motivation for this work comes from the fact that NER is not a task adapted to our problem for two reasons: 
\begin{enumerate}
 \item We do not have a dataset labeled in named entities, which would mean performing a labeling campaign and counting on expert knowledge that we do not have.
 \item Even if we had such a labeled set, a qualitative analysis of the data shows that the presence of a named entity does not correspond to the pathology we have to predict. For example, in the case of suspicions, doubts, or denials, the report may contain a named entity such as ``Apparent \textit{keratosis} is suspected which turns out to be a malignant lump''.
\end{enumerate}
We aim to solve this task, for which we present a hybrid method that combines the transformers with a model based on RoBERTa \cite{carrino-etal-2022-pretrained} and ontologies \cite{10.1093/nar/gkab1063}. To this end, we created intermediate cascade-fashioned models: they detect the type (symptom, neoplastic process, etc.), the anatomical site and severity of the pathology, and a final model that, thanks to the previous ones, predicts the actual pathology. The reports we used come from EHRs and are patient clinical notes written by physicians, which can be a first appointment or a follow-up. Each report has two associated labels: the pathology and a pathology coding. These reports come from the dermatology unit of different health centers in Spain. They have been anonymized semi-automatically with techniques based on symbolic rules. 
Our contributions are as follows: 
\begin{itemize}
 \item An anonymized dataset of dermatology EHRs in Spanish, public and open access.
 \item A typology of dermatological pathologies that enriches existing ontologies and lexicons. 
 \item A hybrid method based on transformers with ontologies for EHRs classification in pathologies.
\end{itemize}
Besides the introduction in Section 1, the article is divided into four sections. Section 2 presents related works where we summarize both similar methods to ours and the linguistic resources that exist. Section 3 provides a description of our methodology and the architecture of the model. Section 4 focuses on 
the results. Section 5 addresses the discussion, conclusions, and possible future work.

\section{Related works}

Text mining in clinical reports has been an important field of NLP for years \cite{mccray1987role}. However, the amount of work related to information extraction in the medical domain has flourished with the expansion of EHRs \cite{pmid24870142}. 
In this regard, in the early 2000s, a combination of ontology and semantic web was often used to extract disease or drug names \cite{lambrix2005towards,Jing2022.05.11.22274984}. In addition, ontologies were used in combination with statistical algorithms \cite{Shannon2021-he}. 

Most of the work in this field is in English, but some, such as \cite{PLN2721,roma2009ontofis}, used Spanish texts. From 2010 onwards, artificial neural networks (NN) such as LSTMs \cite{hochreiter1997long} started to be used, since they have the ability to retain relations between sentences. These NN were used for medical NER in both English \cite{luo2017recurrent} and Spanish \cite{giner2022reconocimiento}. 

Even so, the works that seek to associate the whole text to a pathology concept are scarce in comparison with those that aim to detect the named pathologies.

Regarding English datasets, we find MIMIC \cite{Johnson2023} and MIMIC-III \cite{Johnson2023-fn}. On the other hand, in Spanish there is the private dataset \cite{10.1007/978-3-031-34953-9_38}, SPACCC (\textit{Spanish Clinical Cases Corpus}) which does not contain labeling \cite{intxaurrondo2019spaccc} or PharmacoNER \cite{gonzalez-agirre-etal-2019-pharmaconer} but also for NER. The Spanish dataset closest to ours CodiEsp \cite{miranda2020overview} but is labeled in diagnostic and procedural. 
In order to consult existing datasets, the extensive list created by \cite{10.1093/jamia/ocy173} is a valuable resource. 

In general, as medical data is sensitive, and with the need to comply with the RGPD\footnote{https://www.hacienda.gob.es/es-ES/El\%20Ministerio/Paginas/DPD/Normativa\_PD.aspx}  \cite{9420457}, it is imperative to anonymize clinical data. As explored methods of anonymization, \cite{Lordick2022} use BRAT \cite{stenetorp-etal-2012-brat} to anonymize them. \cite{lima-lopez-etal-2020-hitzalmed} shows that a hybrid model makes it nearly impossible to re-identify individuals. In conjunction with these works, we draw inspiration from \cite{francopoulo:hal-02939437} and MEDOCCAN \cite{marimon2019automatic} to anonymize ours. 

On the other hand, regarding methods that classify the entire text instead of performing NER, we find \cite{krishnan2019ontology}, which employs ontologies to predict diseases in clinical texts. \cite{shoham2023cpllm} applies BERT-inspired transformers \cite{devlin-etal-2019-bert} to create medical word embeddings. The works that provide the best results in terms of accuracy and precision are those that combine the linguistics capacities of Large Language Models (LLMs) and ontologies, so we draw inspiration from these to design our work. For example, \cite{9679070} implemented BERT transformers and medical ontologies, obtaining the best results on MIMIC. To the best of our knowledge, there is no work in Spanish describing how to predict a patient's pathology from the textual EHR using these methods.

\section{Methodology and dataset}


In this section we present the dataset we create and process, as well as the anonymisation technique we use to protect the privacy of its content. Finally, we describe the models we use in addition to our hybrid method: transformer-ontology with cascade models.

\subsection{Dataset description}

\begin{figure*}[h]
 \centering
 \includegraphics[width=\linewidth]{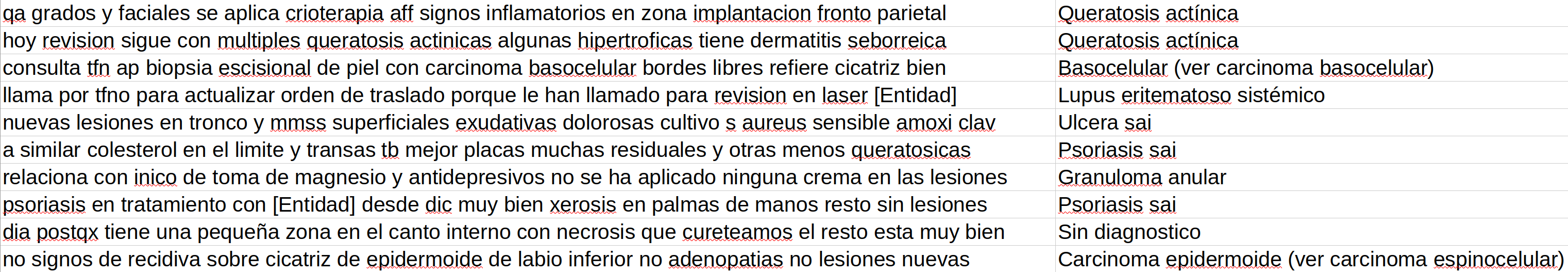}
 \caption{Dataset example. On the left, the first consultation or follow-up report. On the right, the pathology to be predicted.}
 \label{fig_data}
\end{figure*}

The used data correspond to clinical notes from Spanish hospitals with respect to dermatology consultations, both first consultation and subsequent revisions. These data are provided in individual files in HL7 (\textit{Health Level 7}) \cite{Saripalle2019-pz} format, which is a set of international standards that allow the exchange, integration, sharing and retrieval of electronic health data. It also facilitates communication between different systems to be more agile and reliable. 
Our corpus contains 8881 reports and 173 different dermatological pathologies. Each report contains a single label of a dermatological pathology and we are dealing with a case of multiclass classification. The clinical reports are linked to 43 variables of various kinds, including the name and code of the pathology. Given the objective of this project, the dataset has been limited to two variables of interest: the text written by the physician on the consultation in natural language, and the variable that offers the pathology diagnosed to the patient within the taxonomy considered by the collection system. An example of the data set is illustrated in Figure \ref{fig_data}. 

Considering the number of pathologies and their unbalanced distribution (Figure \ref{distr_enf} in Appendix \ref{an_ic}), we hypothesized that even with our hybrid method, many classes would be missed during model training. That is why we studied to define the threshold of the minimum number of examples per pathology that would maximize the accuracy of the model without losing too many pathologies. We decided to keep the 25 most represented pathologies corresponding to a minimum of 61 examples per category. In the appendix \ref{an_umbral}, we explain that this threshold is the optimal threshold to keep a certain number of categories without reducing the efficiency of the models. 

The dataset has not been lemmatized or passed through other classical preprocessing just as neither did \cite{tornberg2023use}, explaining that language models today are capable of processing raw text.

\subsection{Anonymization}

Throughout this section, the process of anonymizing the dataset is detailed. Medical data may contain sensitive information that should not be shared. For this reason, we have proceeded with different semi-automated phases to mask private information. 
At a first level, we proceeded to eliminate all numerical content appearing in these messages. This is because this information is linked to sensitive information: dates, years, ages or different identifiers (ID card, patient identifier, etc.). 
On the other hand, the detection of other types of sensitive information has been addressed, and instead of eliminating them, as has been done with numeric characters, they have been masked with the ``[Entity]” tag. In this case, entities such as proper names, surnames, cities/locations or hospital names have been masked.
To carry out this step, we proceeded with the identification of different external sources to locate such information, using the following:
\begin{itemize}
 \item List of the most frequent male and female surnames and first names in Spain, provided by the INE (National Statistics Institute)\footnote{\url{https://www.ine.es/dyngs/INEbase/es/operacion.htm?c=Estadistica_C&cid=1254736177009&menu=resultados&idp=1254734710990}}.
 \item List of the most frequent words in the Spanish language using resources provided by the Real Academia Española (RAE) linked to the Corpus de Referencia del Español Actual (CREA)\footnote{\url{https://corpus.rae.es/lfrecuencias.html}}.
 \item List of the most common cities and hospitals within the data source used.
\end{itemize}

\begin{figure}[h]
 \centering
 \includegraphics[width=\linewidth]{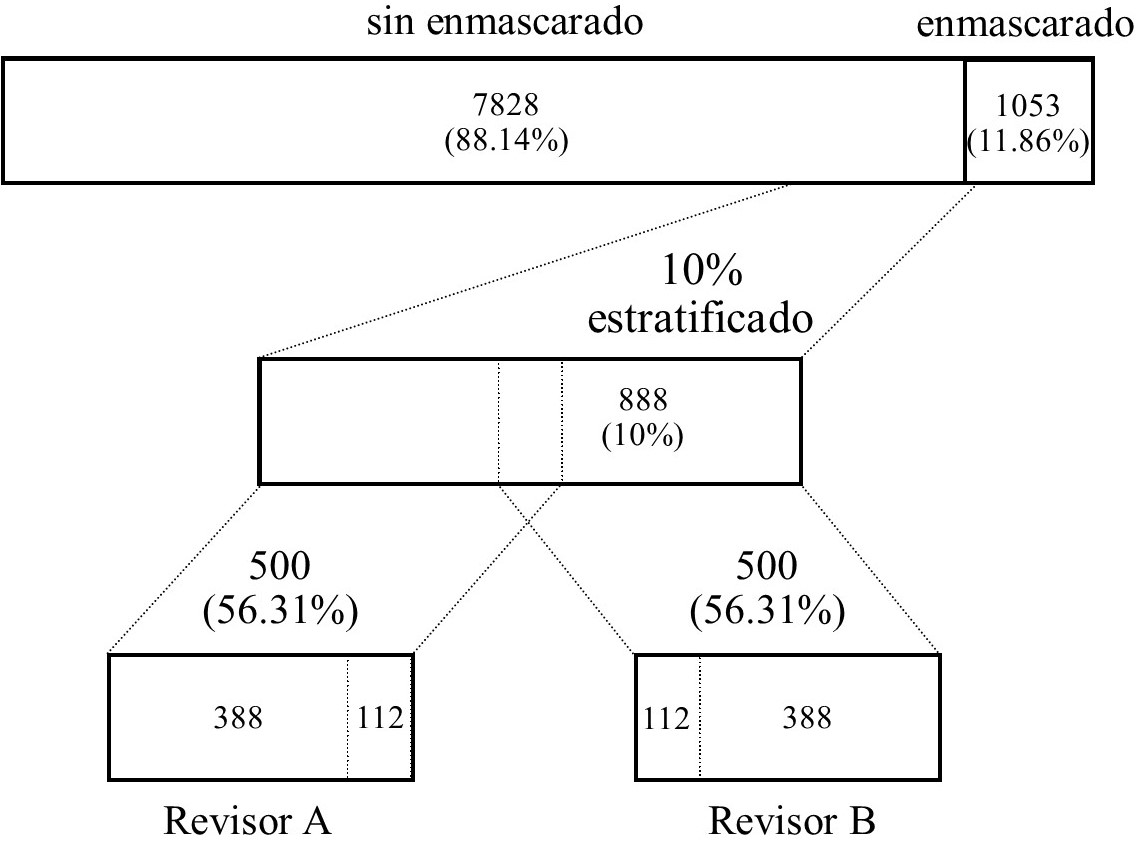}
 \caption{Graphical representation of the partition generated for the validation of the anonymization performed}
 \label{particion}
\end{figure}

Thus, the masking process has been developed as follows:
\begin{itemize}
 \item All those occurrences of the most frequent male or female proper names have been identified.
 \item All those occurrences of the most frequent surnames as first surname have been identified.
 \item Among these names and surnames, those that are among the most frequent terms are filtered so that they are not masked.
 \item A total of 43 exceptions that may be relevant to the particular field of dermatology are added (\textit{cabello}, \textit{seco}, \textit{benigno}, etc.). 
 \item Patterns that have been identified as candidates for containing sensitive information are masked (text following the terms \textit{dr}, \textit{dra}, \textit{doctor}, \textit{doctora}).
\end{itemize}
This generates a set of anonymized and masked texts, which have been analyzed to validate that the information is protected. To this end, a manual review was carried out by two reviewers:
\begin{itemize}
 \item A sample of the anonymized dataset of 10\% of the total has been selected. 
 \item This selection was made in a stratified way with respect to the categorization of the original text. 
 \item This set has been further divided into two subsets of the same size, where each of them has unique entries and entries shared between them.
\end{itemize}
Figure \ref{particion} shows the generated partition, as well as the exact sizes of each of the sets.

Thus, it has been validated whether the anonymization of each text has been adequate or whether it has been mistaken by protecting unnecessary information or not protecting sensitive information.

Following this review, particular cases to be corrected and generalized patterns have been identified. These have been corrected with subsequent analogous iterations to ensure the correct protection of this information. 

At the level of inter-annotator agreement, of the 112 common observations, the reviewers only disagreed on 4. These errors were analyzed and corrected in the previous semi-automatic process for the whole set. 

\subsection{Language Model}

Given the power and amount of data needed to train a language model \cite{kim2024breakthrough}, the most efficient option to be able to use transformers is to perform \textit{fine-tuning}. The goal of this is to find the pre-trained model that best fits our data and the detection of a pathology. For this problem, we chose the bsc-bio-ehr-es \footnote{https://huggingface.co/PlanTL-GOB-ES/bsc-bio-ehr-es}, a pre-trained biomedical-clinical language model designed for the Spanish language. The model has been pre-trained using data from Spanish biomedical and clinical texts to learn specific linguistic patterns. It is based on the RoBERTa \cite{liu2019roberta} architecture. 
However, the results (presented in Section \ref{subs_res}) showed the difficulty of a single model to learn to detect diseases, so we enriched the training using information from medical ontologies and several cascaded models, where each one learns specific information from the data.

\subsection{Ontologies and cascade models}

Throughout this section, we first address the treatment of the imbalance of the target variable, followed by a review of the applied medical and translation ontologies, with a final section focusing on the proposed cascade models.

\subsubsection{Classes imbalance processing}


The imbalance present in the data implies that most classes have near-zero coverage and cause overtraining of the model with the first three classes. 
To remedy this problem, we attempted to reduce the dimensionality not mathematically with PCA \cite{wadud2022can} or T-SNE \cite{liu2021identifying}, but with cascade models, each one trying to solve a simpler task and adding its output to the input of the next model, until being able to predict the exact pathology. 
We were inspired by \cite{dinarelli-rosset-2011-models} and \cite{10.1007/978-3-030-87802-3_19}, who used this method for NER and for speech recognition, respectively.  
On the other hand, instead of using probabilistic methods to reduce the variability of the classes, we introduce determinism in the architecture of our method by means of ontologies that allow us to extract the type of pathology, the anatomical site affected, the severity or the intensity. This allows us to regroup the pathologies into more generic semantic relationships that improve the accuracy in predicting the pathology in each report. 

\subsubsection{Medical ontologies and translation}
 There were previous works that combined machine learning and ontologies \cite{ghidalia2024combining}, but given the linguistic and difficulty characteristics of the task, our method is original in terms of information extraction and combination of specialized models.
Although there are ontologies in the Spanish language such as ONTERMET \cite{vila2015ontoloxias} or ECIEMAPS \cite{villaplana2023improving}, they have the disadvantage of being too specialized or not very complete. For this reason, we decided to automatically translate \cite{stahlberg2020neural} the name of the pathologies in our tag set from Spanish to English with the Google Translate API and to use more general and complete medical ontologies such as UMLS \cite{bodenreider2004unified}, SNOMED \cite{spackman1997snomed}, MedDRA \cite{brown1999medical} and HumanDO \cite{schriml2019human,schriml2022human}.



This information was accessed through the Python libraries \textit{PyMedTermino} and \textit{PyMedTermino2} \cite{Lamy2015-pu}, as well as \textit{medcat} \cite{Kraljevic2021-ln} designed to access these ontologies. With these tools, we identified the coding corresponding to each disease ontology analyzed in a semi-supervised way, checking that the identification matches the actual disease and not possible similar variations. 

By analyzing the features and metadata of these ontologies, we extracted several relevant metadata. First, using SNOMED it is possible to identify different anatomical sites of pathology through the \textit{finding site}. Then, we extracted the type and severity of the disease using UMLS, ICD-10 and MedDRA in combination.
To extract these features we were inspired by the classification of dermatologic predictive features proposed in \cite{Fisher2016-fe}. 

\subsubsection{Cascade models}

The cascade models learn the relationships mentioned in Table \ref{type_transl} (expanded in Table \ref{type_transl_an} in Appendix \ref{an_ic}).

\begin{figure*}[h]
 \centering
 \includegraphics[width=0.7\linewidth]{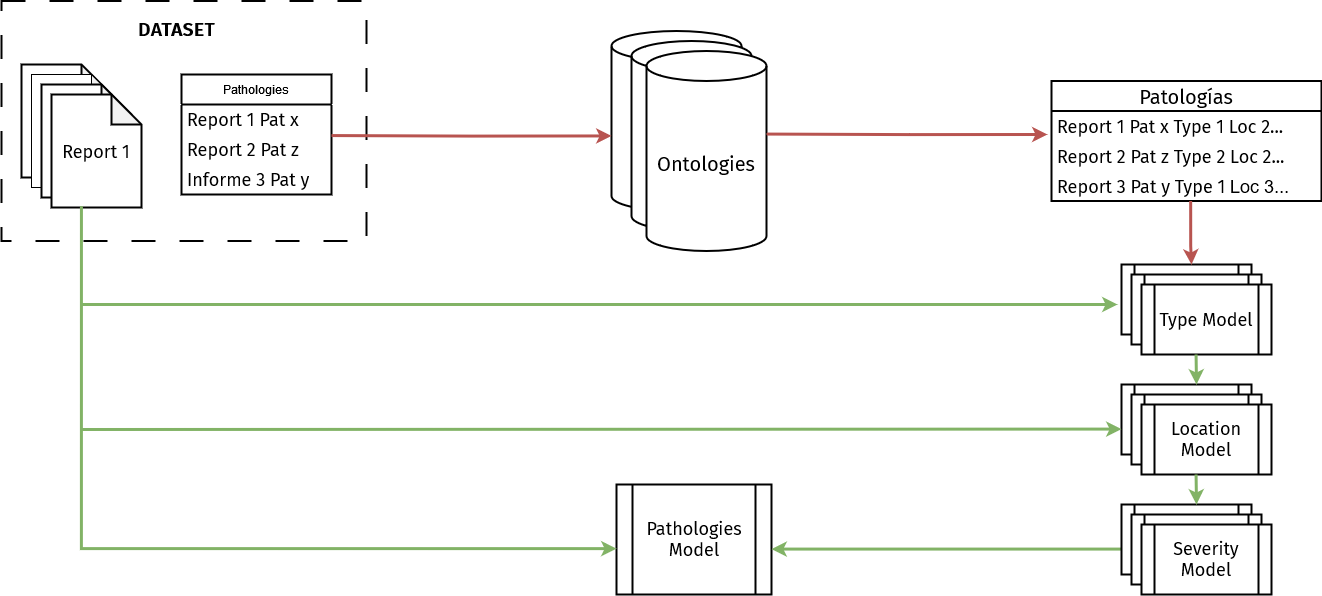}
 \caption{Architecture of our method (in red the training-only stages, in green the training and inference stages)}
 \label{fig_satec}
\end{figure*}

Once we have extracted from the concepts (the translated labels) the common relations existing in the ontology, we train the bsc-bio-ehr-es to predict each of them (see Table \ref{type_transl}). Each predicted relation serves to predict the next one. In Section 4 we show in which order the relationships should be learned. Figure \ref{fig_satec} illustrates our method: 
 
\begin{enumerate}
 \item We extract from the corpus the name of the pathologies, translate them, convert them into concepts and recover the relations linked to these concepts within the ontologies.
 \item For each relation extracted, we train a model. In training mode, the relations are an oracle. In inference mode, the relations are generated by the model.
 \item Each intermediate model receives as input the reports and a relation from the previous model. 
 \item At each stage of the cascade, i.e. when a model has made its prediction, each of them is decoded and concatenated with the initial report, and the predicted report-relationship set is vectorized again using the bsc-bio-ehr-es tokenizer. In the last stage, the input vector contains a representation of the report and the three relationships.
 \item A final model learns to predict the pathology from the reports and the output of the last intermediate model. 
 
\end{enumerate}


\begin{table*}[h]
\begin{center}
\begin{tabular}{|l|l|l|l|}
\hline
Disease & Type & Severity & Location\\
\hline
basal cell carcinoma & neoplastic process & significant & skin \\
  psoriasis & autoimmune process & harmless & extremities \\ 
  melanocytic nevus & precancer & harmless & all \\ 
  acne & disease & mild & all \\ 
  actinic keratosis & precancer & harmless & skin \\ 
\hline
\end{tabular}
\end{center}
\caption{\label{type_transl} Nomenclature generated from the characteristics extracted with \textit{Pymedtermino} (sample of the 5 most frequent diseases, complete version with all diseases and frequencies of occurrence in Table \ref{type_transl_an} in Annex \ref{an_ic})}
\end{table*}

\section{Experiments and results}

In this section we describe the results of our architecture compared to reference models, and evaluate its performance.


\subsection{Enhancement with ontologies}

This strategy consists of training a model prior to the report classification model: it is an intermediate model to learn the type of disease. For this, we used UMLS \cite{bodenreider2004unified} to retrieve the types of pathology, SNOMED for the anatomical site and ICD10 for the severity\footnote{we discarded MedDRA because it did not provide more than the other three}. We automatically translated the name of the diseases with Google Translate. 

\subsection{Models}

Different approaches have been considered for this modeling, both using transformer-based techniques and supervised classification learning models. 

As for those based on transformers, the following have been used:
\begin{itemize}
 \item BETO \cite{CaneteCFP2020}. As indicated by its authors, it is a BERT model trained with a Spanish corpus, using the \textit{Whole Word Masking} technique. 
 \item \textit{bsc-bio-ehr-es} \cite{carrino-etal-2022-pretrained}. Model generated by the BSC (Barcelona Supercomuting Center), using as a basis a large corpus of biomedical texts in Spanish.
\end{itemize}

In addition, the second of these models has been treated through hyperparameter tuning. These hyperparameters and the material used to train the models are specified in Annex \ref{hyper_and_stuff}.
As far as classical machine learning models of classification are concerned, three of the most common algorithms have been used: logistic regression, support vector machines (SVM) and Random Forest. 




Two approaches have been proposed: 
\begin{itemize}
    \item The \textbf{Oracle mode} (OR): the model\footnote{https://huggingface.co/fundacionctic/oracle-dermat} knows the type, the anatomical site and the severity of the pathology. The purpose of this mode is to demonstrate the need for external information to perform the classification task.
    \item The \textbf{Predictive mode} (PR): the final model\footnote{https://huggingface.co/fundacionctic/oracle-predict} must predict the three features mentioned, in the optimal order. Each inference of one feature should help the prediction of the next one. The aim of this way is to demonstrate that our model has a real and useful application for the medical community. 
\end{itemize}

\subsection{Results and evaluation}\label{subs_res}

In order to generate a comparison of OR and PR, we have selected the following metrics: 

\begin{itemize}
    \item Accuracy. Proportion of correct predictions over the total number of predictions made. 
    \item F1-score. Harmonic mean of accuracy and recall. For multi-class problems, the F1-score can be calculated either macro (average of f-score of each class) or micro (average between false positives, false negatives and true positives of the whole set).
    \item Accuracy \textit{top-k}. Proportion of cases in which the true class is among the k most likely predictions of the model. This metric is useful when not only the most likely prediction matters, but also other alternatives that the model considers reasonable. For this entire article, we consider \( k = 2 \).
    
    \item F1-score \textit{top-k}. Extension of the F1-score concept by considering the k most likely classes predicted by the model.
    
\end{itemize}

Drawing inspiration from the work of \cite{NIPS2015_0336dcba,DBLP:journals/corr/abs-2007-15359} on machine learning and \cite{DBLP:journals/corr/cmp-lg-9706014} on the POS-\textit{tagging} task, we consider that medical professionals using our model may find it more informative to have not one but two predictions (looking at the number of possible classes), as this more closely resembles the natural human decision style, and leaving it to the physician to have the final verdict on the disease. 


These metrics have been obtained for all the configurations taken into account in this research. The arrows correspond to the different cascade models: each predicted feature becomes a known feature for the next model.

\subsubsection{Intermediate models results}

A summary of the results obtained with the bsc-bio-ehr-es in PR on each of the three components of the cascade system: type, severity and site of disease is presented in Table \ref{interm-table}. Details of all results can be found in the Appendix \ref{res_inter}. 

\begin{table}[h]
\begin{center}
\resizebox{\linewidth}{!}{
\begin{tabular}{|l|l|l|l|}
\hline
Cat. info. & Acc. &  Micro F1 & Macro F1 \\
\hline
\textbf{t} & 0.57 & 0.56  & 0.38        \\
\textbf{gr} & 0.57 & 0.56  & 0.41     \\
\textbf{sit} & 0.68 & 0.67  & 0.59     \\
\textbf{t} $\rightarrow$ \textbf{gr} $\rightarrow$ \textbf{sit}  & 0.70 & 0.68  & 0.58     \\
\textbf{gr} $\rightarrow$ \textbf{t} $\rightarrow$ \textbf{sit}  & 0.62 & 0.61  & 0.51     \\
\textbf{sit} $\rightarrow$ \textbf{gr} $\rightarrow$ \textbf{t}  & \textbf{0.72} & \textbf{0.71 } &\textbf{0.62}     \\
\hline
\end{tabular}
}
\end{center}
\caption{\label{interm-table} Table with intermediate results for type (\textbf{t}), severity (\textbf{gr}) and site (\textbf{sit}) of each disease}
\end{table}
The best combination of cascade-fashioned intermediate categories seems to be to first predict disease site, followed by severity and finally type. Let us now observe whether this is reflected in the final disease prediction. 

\subsubsection{Final pathology prediction model results}

In all combinations of experiments, either with classical machine learning models or with tuned transformers, this is a single-label multi-class supervised classification task. 

\begin{table*}[h]
\begin{center}
\resizebox{\textwidth}{!}{
\begin{tabular}{|l|l|l|l|l|l|}
\hline
Model & Accuracy & Micro F1-sc. & Macro F1-sc & Acc. \textit{top-k} & F1-sc. \textit{top-k}\\
\hline
Logistic regression (CML) & 0.25 & 0.16 & 0.12 & 0.37 & 0.31\\
SVM (CML) & 0.263 &  0.13  & 0.14 & 0.39 & 0.29\\
Random Forest (CML) & 0.268 & 0.19 & 0.12 & 0.40 & 0.33\\
\hline
PR BETO (TR) & 0.34 & 0.38 & 0.12 & 0.63 & 0.60\\
PR bsc (TR)  & 0.52 & 0.50 & 0.42 & 0.67 & 0.61\\
PR bsc (TR) + \textbf{gr} $\rightarrow$ \textbf{sit} $\rightarrow$ \textbf{t} & 0.58 & 0.55 & 0.47 & 0.69 & 0.62\\
PR bsc (TR) + \textbf{sit} $\rightarrow$ \textbf{gr} $\rightarrow$ \textbf{t} & 0.61 & 0.59 & 0.53 & 0.67 & 0.59\\
PR bsc (TR) + \textbf{t} $\rightarrow$ \textbf{gr} $\rightarrow$ \textbf{sit} & 0.63 & 0.60 & \textbf{0.54} & 0.68 & 0.61\\
PR bsc (TR) + \textbf{t} $\rightarrow$ \textbf{sit} $\rightarrow$ \textbf{gr} & \textbf{0.66} & \textbf{0.61} & 0.38 & \textbf{0.71} & \textbf{0.65}\\
\hline
OR bsc (TR) + \textbf{t} & 0.64 & 0.47 & 0.42 & 0.78 & 0.63\\
OR bsc (TR) + \textbf{gr} & 0.55 & 0.53 & 0.36 & 0.69 & 0.60\\
OR bsc (TR) + \textbf{sit} & 0.65 & 0.63 & 0.50 & 0.75 & 0.74\\
OR bsc (TR) + \textbf{sit} $\rightarrow$ \textbf{t} & 0.77 & 0.76 & 0.66 & 0.87 & 0.87\\
OR bsc (TR) + \textbf{t} $\rightarrow$ \textbf{sit} $\rightarrow$ \textbf{gr} & \textbf{0.84} & \textbf{0.82} & \textbf{0.75} & \textbf{0.92} & \textbf{0.90}\\
\hline
\end{tabular}
}
\end{center}
\caption{\label{tabla_resultados} Table with the metrics obtained for each configuration considered (CML: classical machine learning; TR: transformer; PR predictive mode; OR oracle mode; bsc bsc-bio-ehr-es). }
\end{table*}

Observing Table \ref{tabla_resultados}, it can be seen how the best results with respect to the four metrics are obtained for the model based on ontologies with all the information added, with a substantial difference with the rest of the options. In OR, when the model knows the ontology information, the results exceed 0.84 in absolute accuracy and 0.92 in \textit{top-k} accuracy. In PR, the gain is also significant since we go from 0.5 accuracy with the ``vanilla'' model to 0.66 with the best combination of information. It is remarkable that the best combination of cascaded features is the one of type followed by site and gravity, since gravity has only 4 variables, which made us guess that its learning would be simpler. These results confirm our intuitions: the need to seek external information for training the model and the need to find in which order to learn this information in order to optimize the final classification system, as well as the effectiveness of bsc-bio-ehr-es for our task. The best results in OR and PR for the 5 most frequent pathologies are presented in Table \ref{disease-table}. 

\subsection{Models errors analysis}

\begin{table}[h]
\begin{center}
\resizebox{\linewidth}{!}{
\begin{tabular}{|l|l|l|}
\hline
Disease & F1 PR & F1 OR \\
\hline
acne             & 0.86 & 0.94        \\
basal cell carcin.     & 0.70 & 0.92        \\
psoriasis           & 0.81 & 0.87        \\
melanocytic nevus       & 0.72 & 0.93        \\
actinic queratosis       & 0.63 & 0.83        \\

\hline
\end{tabular}
}
\end{center}
\caption{\label{disease-table} Table with the metrics (without \textit{top-k}) for the 5 most frequent pathologies}
\end{table}

After the generation of these models, an error analysis of the best PR configuration and the best OR was performed. 

On the one hand, the poor accuracy of the three classical supervised learning models (logistic regression, SVM, Random Forest) has as a possible cause the inability to generalize over infrequent labels.
The main limitation of the classical machine learning models lies in their lack of context memory: the longer a text is, the more costly it is for the machine to remember the content of the beginning. The high dimensions of embeddings do not help either, since the number of variables to be learned is exponential.

Examining the disease categories where the models err most, we conclude that the discrepancies are due to the fact that the confused diseases share similar affected body areas, similar severity levels, and some shared visual appearance and symptom descriptions. On the other hand, with the exception of cancers, the model tends to confound diseases whose physical appearance is lumps.
The most frequent confusion is between basal cell and squamous cell carcinoma, accounting for 844 of 2334 errors. Although they can appear anywhere on the body, these diseases are more likely to develop in sun-exposed areas, such as the head, neck and arms. The key difference between them is their severity, with squamous cell carcinoma being more aggressive.
Another frequent confusion of the model is acne with seborrheic keratosis with 325 errors. 
These diseases have similarities in visual appearance (bumps, redness, itching) and in sites of appearance (face and torso). These are two dermatological conditions that may resemble each other in descriptive texts, which could lead to confusion in a language model.


\section{Discussion, conclusion and future works}

Generally speaking, the errors made by the model can be explained, despite the difficulty of the task (many possible diseases, medical reports that can be either first appointment or follow-up). \newline
Because of these errors we have decided to use a metric based on \textit{top-k} rather than strict accuracy. If this system is used in a medical setting, we understand that a practitioner would prefer to have to choose between 2 possible diagnoses rather than 25. 
It is also interesting to mention that the best scenario in PR is when the model has to learn both the anatomical site and the type and severity of the disease before guessing the latter. We conclude that a transformer needs such external information to perform classification because its language model is not sufficient. At first, we guessed that the model that first learns the severity would give the best result, since this variable has only 4 categories (harmless, mild, moderate, extreme). However, the best result is given by a model that first learns the type of disease, followed by the location and finally the severity of the disease. That means, and it makes medical sense, that the type of pathology is detected first, before trying to guess its severity. \newline
As conclusions, our work presents several important contributions. First, we provide a new Spanish dermatology EHR dataset, anonymized and labeled in pathology. Second, we propose a novel method to predict in EHR the pathology that a person suffers from. 
Third, this method is a hybrid system composed of several cascaded specialized transforming models that use as input, in addition to the EHRs, the output of the preceding model. The last model is the one that predicts the exact pathology. 
The results show that cascading models are more effective than a single model in distinguishing rare diseases. That means that a single end-to-end model of transformers is not sufficient to distinguish one pathology concept among several options in an EHR, but also that the use of external ontologies is necessary for transformers to learn pathology-related intermediate concepts, similar to human learning. Still, the results show significant room for improvement. This can be approached from different points of view, such as manual labeling of EHRs at first or follow-up appointment, finding new ontology relations (such as more exhaustive etiology features), or trying different models to learn each feature. \\
As a possible line of future work we propose to automate the use of ontologies using the RAG \cite{gao2024retrievalaugmented}, as in BiomedRAG \cite{li2024biomedrag} as well as the use of NegEx \cite{Arguello-Gonzalez2023} to avoid false positives, accompanied by our cascade modeling system to eliminate determinism when finding concept relationships to become an end-to-end model. 

\section*{Acknowledgements}

Thanks to SATEC for providing the dataset, to the CTIC Foundation for making all the material and resources available to us. We thank and congratulate the BSC for the bsc-bio-ehr-es model.

\bibliographystyle{fullname_esp}
\bibliography{EjemploARTsepln}

\begin{thebibliography}{}

\bibitem[\protect\citename{Aberkane, Poels, y Broucke}2021]{9420457}
Aberkane, A.-J., G.~Poels, y S.~V. Broucke.
\newblock 2021.
\newblock Exploring automated gdpr-compliance in requirements engineering: A systematic mapping study.
\newblock {\em IEEE Access}, 9:66542--66559.

\bibitem[\protect\citename{Aracena \bgroup et al.\egroup }2023]{aracena-etal-2023-pre}
Aracena, C., N.~Rodr{\'\i}guez, V.~Rocco, y J.~Dunstan.
\newblock 2023.
\newblock Pre-trained language models in {S}panish for health insurance coverage.
\newblock En T.~Naumann A.~Ben~Abacha S.~Bethard K.~Roberts, y A.~Rumshisky, editores, {\em Proceedings of the 5th Clinical Natural Language Processing Workshop}, p\'aginas 433--438, Toronto, Canada, Julio. Association for Computational Linguistics.

\bibitem[\protect\citename{Araki \bgroup et al.\egroup }2023]{Araki2023}
Araki, K., N.~Matsumoto, K.~Togo, N.~Yonemoto, E.~Ohki, L.~Xu, Y.~Hasegawa, D.~Satoh, R.~Takemoto, y T.~Miyazaki.
\newblock 2023.
\newblock Developing artificial intelligence models for extracting oncologic outcomes from japanese electronic health records.
\newblock {\em Advances in Therapy}, Mar.

\bibitem[\protect\citename{Arg{\"u}ello-Gonz{\'a}lez \bgroup et al.\egroup }2023]{Arguello-Gonzalez2023}
Arg{\"u}ello-Gonz{\'a}lez, G., J.~Aquino-Esperanza, D.~Salvador, R.~Bret{\'o}n-Romero, C.~Del R{\'i}o-Bermudez, J.~Tello, y S.~Menke.
\newblock 2023.
\newblock Negation recognition in clinical natural language processing using a combination of the negex algorithm and a convolutional neural network.
\newblock {\em BMC Medical Informatics and Decision Making}, 23(1):216, Oct.

\bibitem[\protect\citename{Bodenreider}2004]{bodenreider2004unified}
Bodenreider, O.
\newblock 2004.
\newblock The unified medical language system (umls): integrating biomedical terminology.
\newblock {\em Nucleic acids research}, 32(suppl\_1):D267--D270.

\bibitem[\protect\citename{Brown, Wood, y Wood}1999]{brown1999medical}
Brown, E.~G., L.~Wood, y S.~Wood.
\newblock 1999.
\newblock The medical dictionary for regulatory activities (meddra).
\newblock {\em Drug safety}, 20(2):109--117.

\bibitem[\protect\citename{Buchan, Filannino, y Özlem Uzuner}2017]{BUCHAN201723}
Buchan, K., M.~Filannino, y Özlem Uzuner.
\newblock 2017.
\newblock Automatic prediction of coronary artery disease from clinical narratives.
\newblock {\em Journal of Biomedical Informatics}, 72:23--32.

\bibitem[\protect\citename{Carrino \bgroup et al.\egroup }2022]{carrino-etal-2022-pretrained}
Carrino, C.~P., J.~Llop, M.~P{\`a}mies, A.~Guti{\'e}rrez-Fandi{\~n}o, J.~Armengol-Estap{\'e}, J.~Silveira-Ocampo, A.~Valencia, A.~Gonzalez-Agirre, y M.~Villegas.
\newblock 2022.
\newblock Pretrained biomedical language models for clinical {NLP} in {S}panish.
\newblock En D.~Demner-Fushman K.~B. Cohen S.~Ananiadou, y J.~Tsujii, editores, {\em Proceedings of the 21st Workshop on Biomedical Language Processing}, p\'aginas 193--199, Dublin, Ireland, Mayo. Association for Computational Linguistics.

\bibitem[\protect\citename{Cañete \bgroup et al.\egroup }2020]{CaneteCFP2020}
Cañete, J., G.~Chaperon, R.~Fuentes, J.-H. Ho, H.~Kang, y J.~Pérez.
\newblock 2020.
\newblock Spanish pre-trained bert model and evaluation data.
\newblock En {\em PML4DC at ICLR 2020}.

\bibitem[\protect\citename{Chizhikova \bgroup et al.\egroup }2023]{CHIZHIKOVA2023106581}
Chizhikova, M., P.~López-Úbeda, J.~Collado-Montañez, T.~Martín-Noguerol, M.~C. Díaz-Galiano, A.~Luna, L.~A. Ureña-López, y M.~T. Martín-Valdivia.
\newblock 2023.
\newblock Cares: A corpus for classification of spanish radiological reports.
\newblock {\em Computers in Biology and Medicine}, 154:106581.

\bibitem[\protect\citename{De~Freitas \bgroup et al.\egroup }2021]{DeFreitas2021}
De~Freitas, J.~K., K.~W. Johnson, E.~Golden, G.~N. Nadkarni, J.~T. Dudley, E.~P. Bottinger, B.~S. Glicksberg, y R.~Miotto.
\newblock 2021.
\newblock Phe2vec: Automated disease phenotyping based on unsupervised embeddings from electronic health records.
\newblock {\em Patterns}, Sep.

\bibitem[\protect\citename{de~la Rosa \bgroup et al.\egroup }2022]{PLN6403}
de~la Rosa, J., E.~G. Ponferrada, M.~Romero, P.~Villegas, P.~G. de~Prado~Salas, y M.~Grandury.
\newblock 2022.
\newblock Bertin: Efficient pre-training of a spanish language model using perplexity sampling.
\newblock {\em Procesamiento del Lenguaje Natural}, 68(0):13--23.

\bibitem[\protect\citename{Devlin \bgroup et al.\egroup }2019]{devlin-etal-2019-bert}
Devlin, J., M.-W. Chang, K.~Lee, y K.~Toutanova.
\newblock 2019.
\newblock {BERT}: Pre-training of deep bidirectional transformers for language understanding.
\newblock En J.~Burstein C.~Doran, y T.~Solorio, editores, {\em Proceedings of the 2019 Conference of the North {A}merican Chapter of the Association for Computational Linguistics: Human Language Technologies, Volume 1 (Long and Short Papers)}, p\'aginas 4171--4186, Minneapolis, Minnesota, Junio. Association for Computational Linguistics.

\bibitem[\protect\citename{Dinarelli y Rosset}2011]{dinarelli-rosset-2011-models}
Dinarelli, M. y S.~Rosset.
\newblock 2011.
\newblock Models cascade for tree-structured named entity detection.
\newblock En H.~Wang y D.~Yarowsky, editores, {\em Proceedings of 5th International Joint Conference on Natural Language Processing}, p\'aginas 1269--1278, Chiang Mai, Thailand, Noviembre. Asian Federation of Natural Language Processing.

\bibitem[\protect\citename{Doan \bgroup et al.\egroup }2014]{pmid24870142}
Doan, S., M.~Conway, T.~M. Phuong, y L.~Ohno-Machado.
\newblock 2014.
\newblock {{N}atural language processing in biomedicine: a unified system architecture overview}.
\newblock {\em Methods Mol Biol}, 1168:275--294.

\bibitem[\protect\citename{Fisher \bgroup et al.\egroup }2016]{Fisher2016-fe}
Fisher, H.~M., R.~Hoehndorf, B.~S. Bazelato, S.~S. Dadras, L.~E. King, Jr, G.~V. Gkoutos, J.~P. Sundberg, y P.~N. Schofield.
\newblock 2016.
\newblock {DermO}; an ontology for the description of dermatologic disease.
\newblock {\em J. Biomed. Semantics}, 7(1):38.

\bibitem[\protect\citename{Francopoulo y Schaub}2020]{francopoulo:hal-02939437}
Francopoulo, G. y L.-P. Schaub.
\newblock 2020.
\newblock {Anonymization for the GDPR in the Context of Citizen and Customer Relationship Management and NLP}.
\newblock En {\em {workshop on Legal and Ethical Issues (Legal2020)}}, p\'aginas 9--14, Marseille, France, Mayo. {LREC2020}, {ELRA}.

\bibitem[\protect\citename{Gao \bgroup et al.\egroup }2024]{gao2024retrievalaugmented}
Gao, Y., Y.~Xiong, X.~Gao, K.~Jia, J.~Pan, Y.~Bi, Y.~Dai, J.~Sun, M.~Wang, y H.~Wang.
\newblock 2024.
\newblock Retrieval-augmented generation for large language models: A survey.

\bibitem[\protect\citename{García \bgroup et al.\egroup }2007]{PLN2721}
García, F.~C., J.~M.~G. Hidalgo, M.~de~Buenaga~Rodríguez, J.~Mata, y M.~M. López.
\newblock 2007.
\newblock Acceso a la información bilingue utilizando ontologías específicas del dominio biomédico.
\newblock {\em Procesamiento del Lenguaje Natural}, 38(0).

\bibitem[\protect\citename{Ghannay \bgroup et al.\egroup }2021]{10.1007/978-3-030-87802-3_19}
Ghannay, S., A.~Caubri\`{e}re, S.~Mdhaffar, G.~Laperri\`{e}re, B.~Jabaian, y Y.~Est\`{e}ve.
\newblock 2021.
\newblock Where are we in semantic concept extraction for spoken language understanding?
\newblock En {\em Speech and Computer: 23rd International Conference, SPECOM 2021, St. Petersburg, Russia, September 27–30, 2021, Proceedings}, p\'agina 202–213, Berlin, Heidelberg. Springer-Verlag.

\bibitem[\protect\citename{Ghidalia \bgroup et al.\egroup }2024]{ghidalia2024combining}
Ghidalia, S., O.~L. Narsis, A.~Bertaux, y C.~Nicolle.
\newblock 2024.
\newblock Combining machine learning and ontology: A systematic literature review.

\bibitem[\protect\citename{Giner P{\'e}rez~de Luc{\'\i}a}2022]{giner2022reconocimiento}
Giner P{\'e}rez~de Luc{\'\i}a, J.
\newblock 2022.
\newblock {\em Reconocimiento de entidades nombradas mediante t{\'e}cnicas de aprendizaje neuronal profundo en im{\'a}genes manuscritas}.
\newblock {Ph.D.} tesis, Universitat Polit{\`e}cnica de Val{\`e}ncia.

\bibitem[\protect\citename{Gonzalez-Agirre \bgroup et al.\egroup }2019]{gonzalez-agirre-etal-2019-pharmaconer}
Gonzalez-Agirre, A., M.~Marimon, A.~Intxaurrondo, O.~Rabal, M.~Villegas, y M.~Krallinger.
\newblock 2019.
\newblock {P}harma{C}o{NER}: Pharmacological substances, compounds and proteins named entity recognition track.
\newblock En K.~Jin-Dong N.~Claire B.~Robert, y D.~Louise, editores, {\em Proceedings of the 5th Workshop on BioNLP Open Shared Tasks}, p\'aginas 1--10, Hong Kong, China, Noviembre. Association for Computational Linguistics.

\bibitem[\protect\citename{Haug \bgroup et al.\egroup }2013]{10.1136/amiajnl-2012-001376}
Haug, P.~J., J.~P. Ferraro, J.~Holmen, X.~Wu, K.~Mynam, M.~Ebert, N.~Dean, y J.~Jones.
\newblock 2013.
\newblock {An ontology-driven, diagnostic modeling system}.
\newblock {\em Journal of the American Medical Informatics Association}, 20(e1):e102--e110, 03.

\bibitem[\protect\citename{Hochreiter y Schmidhuber}1997]{hochreiter1997long}
Hochreiter, S. y J.~Schmidhuber.
\newblock 1997.
\newblock Long short-term memory.
\newblock {\em Neural computation}, 9(8):1735--1780.

\bibitem[\protect\citename{Intxaurrondo}2019]{intxaurrondo2019spaccc}
Intxaurrondo, A.
\newblock 2019.
\newblock Spaccc (spanish clinical case corpus) tokenizer.

\bibitem[\protect\citename{Jing \bgroup et al.\egroup }2022]{Jing2022.05.11.22274984}
Jing, X., H.~Min, Y.~Gong, D.~F. Sittig, P.~Biondich, D.~Robinson, T.~Law, A.~Wright, C.~N{\o}hr, A.~Faxvaag, L.~Rennert, N.~Hubig, y R.~Gimbel.
\newblock 2022.
\newblock A systematic review of ontology-based clinical decision support system rules: usage, management, and interoperability.
\newblock {\em medRxiv}.

\bibitem[\protect\citename{Johnson, Pollard, y Mark}2023]{Johnson2023-fn}
Johnson, A., T.~Pollard, y R.~Mark.
\newblock 2023.
\newblock {MIMIC-III} clinical database.

\bibitem[\protect\citename{Johnson \bgroup et al.\egroup }2023]{Johnson2023}
Johnson, A. E.~W., L.~Bulgarelli, L.~Shen, A.~Gayles, A.~Shammout, S.~Horng, T.~J. Pollard, S.~Hao, B.~Moody, B.~Gow, L.-w.~H. Lehman, L.~A. Celi, y R.~G. Mark.
\newblock 2023.
\newblock Mimic-iv, a freely accessible electronic health record dataset.
\newblock {\em Scientific Data}, 10(1):1, Jan.

\bibitem[\protect\citename{Kim \bgroup et al.\egroup }2024]{kim2024breakthrough}
Kim, B., S.~Cha, S.~Park, J.~Lee, S.~Lee, S.-h. Kang, J.~So, K.~Kim, J.~Jung, J.-G. Lee, et~al.
\newblock 2024.
\newblock The breakthrough memory solutions for improved performance on llm inference.
\newblock {\em IEEE Micro}.

\bibitem[\protect\citename{Koleck \bgroup et al.\egroup }2019]{10.1093/jamia/ocy173}
Koleck, T.~A., C.~Dreisbach, P.~E. Bourne, y S.~Bakken.
\newblock 2019.
\newblock {Natural language processing of symptoms documented in free-text narratives of electronic health records: a systematic review}.
\newblock {\em Journal of the American Medical Informatics Association}, 26(4):364--379, 02.

\bibitem[\protect\citename{Kraljevic \bgroup et al.\egroup }2021a]{Kraljevic2021-ln}
Kraljevic, Z., T.~Searle, A.~Shek, L.~Roguski, K.~Noor, D.~Bean, A.~Mascio, L.~Zhu, A.~A. Folarin, A.~Roberts, R.~Bendayan, M.~P. Richardson, R.~Stewart, A.~D. Shah, W.~K. Wong, Z.~Ibrahim, J.~T. Teo, y R.~J.~B. Dobson.
\newblock 2021a.
\newblock Multi-domain clinical natural language processing with {MedCAT}: The medical concept annotation toolkit.
\newblock {\em Artif. Intell. Med.}, 117:102083, Julio.

\bibitem[\protect\citename{Kraljevic \bgroup et al.\egroup }2021b]{DBLPjournals/corr/abs-2107-03134}
Kraljevic, Z., A.~Shek, D.~Bean, R.~Bendayan, J.~T. Teo, y R.~J.~B. Dobson.
\newblock 2021b.
\newblock Medgpt: Medical concept prediction from clinical narratives.
\newblock {\em CoRR}, abs/2107.03134.

\bibitem[\protect\citename{Krishnan y Kamath~S}2019]{krishnan2019ontology}
Krishnan, G.~S. y S.~Kamath~S.
\newblock 2019.
\newblock Ontology-driven text feature modeling for disease prediction using unstructured radiological notes.
\newblock {\em Computaci{\'o}n y Sistemas}, 23(3):915--922.

\bibitem[\protect\citename{Lambrix}2005]{lambrix2005towards}
Lambrix, P.
\newblock 2005.
\newblock Towards a semantic web for bioinformatics using ontology-based annotation.
\newblock En {\em 14th IEEE International Workshops on Enabling Technologies: Infrastructure for Collaborative Enterprise (WETICE'05)}, p\'aginas 3--7. IEEE.

\bibitem[\protect\citename{Lamy, Venot, y Duclos}2015]{Lamy2015-pu}
Lamy, J.-B., A.~Venot, y C.~Duclos.
\newblock 2015.
\newblock {PyMedTermino}: an open-source generic {API} for advanced terminology services.
\newblock {\em Stud. Health Technol. Inform.}, 210:924--928.

\bibitem[\protect\citename{Lapin, Hein, y Schiele}2015]{NIPS2015_0336dcba}
Lapin, M., M.~Hein, y B.~Schiele.
\newblock 2015.
\newblock Top-k multiclass svm.
\newblock En C.~Cortes N.~Lawrence D.~Lee M.~Sugiyama, y R.~Garnett, editores, {\em Advances in Neural Information Processing Systems}, volumen~28. Curran Associates, Inc.

\bibitem[\protect\citename{Le \bgroup et al.\egroup }2022]{9903583}
Le, T.-D., R.~Noumeir, J.~Rambaud, G.~Sans, y P.~Jouvet.
\newblock 2022.
\newblock Detecting of a patient's condition from clinical narratives using natural language representation.
\newblock {\em IEEE Open Journal of Engineering in Medicine and Biology}, 3:142--149.

\bibitem[\protect\citename{Li \bgroup et al.\egroup }2024]{li2024biomedrag}
Li, M., H.~Kilicoglu, H.~Xu, y R.~Zhang.
\newblock 2024.
\newblock Biomedrag: A retrieval augmented large language model for biomedicine.

\bibitem[\protect\citename{Li \bgroup et al.\egroup }2023]{9964038}
Li, Y., M.~Mamouei, G.~Salimi-Khorshidi, S.~Rao, A.~Hassaine, D.~Canoy, T.~Lukasiewicz, y K.~Rahimi.
\newblock 2023.
\newblock Hi-behrt: Hierarchical transformer-based model for accurate prediction of clinical events using multimodal longitudinal electronic health records.
\newblock {\em IEEE Journal of Biomedical and Health Informatics}, 27(2):1106--1117.

\bibitem[\protect\citename{Lima~Lopez \bgroup et al.\egroup }2020]{lima-lopez-etal-2020-hitzalmed}
Lima~Lopez, S., N.~Perez, L.~Garc{\'\i}a-Sardi{\~n}a, y M.~Cuadros.
\newblock 2020.
\newblock {H}itzal{M}ed: Anonymisation of clinical text in {S}panish.
\newblock En N.~Calzolari F.~B{\'e}chet P.~Blache K.~Choukri C.~Cieri T.~Declerck S.~Goggi H.~Isahara B.~Maegaard J.~Mariani H.~Mazo A.~Moreno J.~Odijk, y S.~Piperidis, editores, {\em Proceedings of the Twelfth Language Resources and Evaluation Conference}, p\'aginas 7038--7043, Marseille, France, Mayo. European Language Resources Association.

\bibitem[\protect\citename{Liu \bgroup et al.\egroup }2021]{liu2021identifying}
Liu, G., M.~Boyd, M.~Yu, S.~Z. Halim, y N.~Quddus.
\newblock 2021.
\newblock Identifying causality and contributory factors of pipeline incidents by employing natural language processing and text mining techniques.
\newblock {\em Process safety and environmental protection}, 152:37--46.

\bibitem[\protect\citename{Liu \bgroup et al.\egroup }2019]{liu2019roberta}
Liu, Y., M.~Ott, N.~Goyal, J.~Du, M.~Joshi, D.~Chen, O.~Levy, M.~Lewis, L.~Zettlemoyer, y V.~Stoyanov.
\newblock 2019.
\newblock Roberta: A robustly optimized bert pretraining approach.
\newblock {\em arXiv preprint arXiv:1907.11692}.

\bibitem[\protect\citename{Lordick, Hoch, y Fransen}2022]{Lordick2022}
Lordick, T., A.~Hoch, y B.~Fransen, 2022.
\newblock {\em Anonymization of Electronic Health Care Records: The EHR Anonymizer}, p\'aginas 485--499.
\newblock Springer International Publishing, Cham.

\bibitem[\protect\citename{Luo}2017]{luo2017recurrent}
Luo, Y.
\newblock 2017.
\newblock Recurrent neural networks for classifying relations in clinical notes.
\newblock {\em Journal of biomedical informatics}, 72:85--95.

\bibitem[\protect\citename{Marimon \bgroup et al.\egroup }2019]{marimon2019automatic}
Marimon, M., A.~Gonzalez-Agirre, A.~Intxaurrondo, H.~Rodriguez, J.~L. Martin, M.~Villegas, y M.~Krallinger.
\newblock 2019.
\newblock Automatic de-identification of medical texts in spanish: the meddocan track, corpus, guidelines, methods and evaluation of results.
\newblock En {\em IberLEF@ SEPLN}, p\'aginas 618--638.

\bibitem[\protect\citename{McCray \bgroup et al.\egroup }1987]{mccray1987role}
McCray, A.~T., J.~L. Sponsler, B.~Brylawski, y A.~C. Browne.
\newblock 1987.
\newblock The role of lexical knowledge in biomedical text understanding.
\newblock En {\em proceedings of the annual symposium on computer application in medical care}, p\'agina 103. American Medical Informatics Association.

\bibitem[\protect\citename{Miranda-Escalada \bgroup et al.\egroup }2020]{miranda2020overview}
Miranda-Escalada, A., A.~Gonzalez-Agirre, J.~Armengol-Estap{\'e}, y M.~Krallinger.
\newblock 2020.
\newblock Overview of automatic clinical coding: annotations, guidelines, and solutions for non-english clinical cases at codiesp track of clef ehealth 2020.
\newblock En {\em Working Notes of Conference and Labs of the Evaluation (CLEF) Forum. CEUR Workshop Proceedings}.

\bibitem[\protect\citename{Moreno-Barea \bgroup et al.\egroup }2023]{10.1007/978-3-031-34953-9_38}
Moreno-Barea, F.~J., H.~Mesa, N.~Ribelles, E.~Alba, y J.~M. Jerez.
\newblock 2023.
\newblock Clinical text classification in cancer real-world data in spanish.
\newblock En I.~Rojas O.~Valenzuela F.~Rojas~Ruiz L.~J. Herrera, y F.~Ortu{\~{n}}o, editores, {\em Bioinformatics and Biomedical Engineering}, p\'aginas 482--496, Cham. Springer Nature Switzerland.

\bibitem[\protect\citename{N{\'e}v{\'e}ol \bgroup et al.\egroup }2018]{Neveol2018}
N{\'e}v{\'e}ol, A., H.~Dalianis, S.~Velupillai, G.~Savova, y P.~Zweigenbaum.
\newblock 2018.
\newblock Clinical natural language processing in languages other than english: opportunities and challenges.
\newblock {\em Journal of Biomedical Semantics}, 9(1):12, Mar.

\bibitem[\protect\citename{Otter, Medina, y Kalita}2021]{9075398}
Otter, D.~W., J.~R. Medina, y J.~K. Kalita.
\newblock 2021.
\newblock A survey of the usages of deep learning for natural language processing.
\newblock {\em IEEE Transactions on Neural Networks and Learning Systems}, 32(2):604--624.

\bibitem[\protect\citename{Peng \bgroup et al.\egroup }2021]{9679070}
Peng, X., G.~Long, T.~Shen, S.~Wang, y J.~Jiang.
\newblock 2021.
\newblock Sequential diagnosis prediction with transformer and ontological representation.
\newblock En {\em 2021 IEEE International Conference on Data Mining (ICDM)}, p\'aginas 489--498, Los Alamitos, CA, USA, dec. IEEE Computer Society.

\bibitem[\protect\citename{Ratnaparkhi}1997]{DBLP:journals/corr/cmp-lg-9706014}
Ratnaparkhi, A.
\newblock 1997.
\newblock A linear observed time statistical parser based on maximum entropy models.
\newblock {\em CoRR}, cmp-lg/9706014.

\bibitem[\protect\citename{Rojas, Dunstan, y Villena}2022]{rojas-etal-2022-clinical}
Rojas, M., J.~Dunstan, y F.~Villena.
\newblock 2022.
\newblock Clinical flair: A pre-trained language model for {S}panish clinical natural language processing.
\newblock En T.~Naumann S.~Bethard K.~Roberts, y A.~Rumshisky, editores, {\em Proceedings of the 4th Clinical Natural Language Processing Workshop}, p\'aginas 87--92, Seattle, WA, Julio. Association for Computational Linguistics.

\bibitem[\protect\citename{Rom{\'a}-Ferri}2009]{roma2009ontofis}
Rom{\'a}-Ferri, M.~T.
\newblock 2009.
\newblock {\em OntoFIS: tecnolog{\'\i}a ontol{\'o}gica en el dominio farmacoterap{\'e}utico}.
\newblock Universidad de Alicante.

\bibitem[\protect\citename{Saripalle, Runyan, y Russell}2019]{Saripalle2019-pz}
Saripalle, R., C.~Runyan, y M.~Russell.
\newblock 2019.
\newblock Using {HL7} {FHIR} to achieve interoperability in patient health record.
\newblock {\em J. Biomed. Inform.}, 94(103188):103188.

\bibitem[\protect\citename{Sawada, Kaneko, y Sagi}2020]{DBLP:journals/corr/abs-2007-15359}
Sawada, A., E.~Kaneko, y K.~Sagi.
\newblock 2020.
\newblock Trade-offs in top-k classification accuracies on losses for deep learning.
\newblock {\em CoRR}, abs/2007.15359.

\bibitem[\protect\citename{Schaub y Mir{\'o}~Maestre}2023]{schaub2023inteligencia}
Schaub, L.~P. y M.~Mir{\'o}~Maestre.
\newblock 2023.
\newblock La inteligencia artificial como impulso del procesamiento del lenguaje natural: retos, fronteras y logros.
\newblock {\em Abaco: Revista de cultura y ciencias sociales}, 61(118):66--81.

\bibitem[\protect\citename{Schriml \bgroup et al.\egroup }2019]{schriml2019human}
Schriml, L.~M., E.~Mitraka, J.~Munro, B.~Tauber, M.~Schor, L.~Nickle, V.~Felix, L.~Jeng, C.~Bearer, R.~Lichenstein, y others.
\newblock 2019.
\newblock Human disease ontology 2018 update: classification, content and workflow expansion.
\newblock {\em Nucleic acids research}, 47(D1):D955--D962.

\bibitem[\protect\citename{Schriml \bgroup et al.\egroup }2022]{schriml2022human}
Schriml, L.~M., J.~B. Munro, M.~Schor, D.~Olley, C.~McCracken, V.~Felix, J.~A. Baron, R.~Jackson, S.~M. Bello, C.~Bearer, et~al.
\newblock 2022.
\newblock The human disease ontology 2022 update.
\newblock {\em Nucleic acids research}, 50(D1):D1255--D1261.

\bibitem[\protect\citename{Schriml \bgroup et al.\egroup }2021]{10.1093/nar/gkab1063}
Schriml, L.~M., J.~B. Munro, M.~Schor, D.~Olley, C.~McCracken, V.~Felix, J.~Baron, R.~Jackson, S.~Bello, C.~Bearer, R.~Lichenstein, K.~Bisordi, N.~C. Dialo, M.~Giglio, y C.~Greene.
\newblock 2021.
\newblock {The Human Disease Ontology 2022 update}.
\newblock {\em Nucleic Acids Research}, 50(D1):D1255--D1261, 11.

\bibitem[\protect\citename{Shannon \bgroup et al.\egroup }2021]{Shannon2021-he}
Shannon, G.~J., N.~Rayapati, S.~M. Corns, y D.~C. Wunsch, 2nd.
\newblock 2021.
\newblock Comparative study using inverse ontology cogency and alternatives for concept recognition in the annotated national library of medicine database.
\newblock {\em Neural Netw.}, 139:86--104, Julio.

\bibitem[\protect\citename{Shoham y Rappoport}2023]{shoham2023cpllm}
Shoham, O.~B. y N.~Rappoport.
\newblock 2023.
\newblock Cpllm: Clinical prediction with large language models.

\bibitem[\protect\citename{Spackman, Campbell, y C{\^o}t{\'e}}1997]{spackman1997snomed}
Spackman, K.~A., K.~E. Campbell, y R.~A. C{\^o}t{\'e}.
\newblock 1997.
\newblock Snomed rt: a reference terminology for health care.
\newblock En {\em Proceedings of the AMIA annual fall symposium}, p\'agina 640. American Medical Informatics Association.

\bibitem[\protect\citename{Stahlberg}2020]{stahlberg2020neural}
Stahlberg, F.
\newblock 2020.
\newblock Neural machine translation: A review.
\newblock {\em Journal of Artificial Intelligence Research}, 69:343--418.

\bibitem[\protect\citename{Stenetorp \bgroup et al.\egroup }2012]{stenetorp-etal-2012-brat}
Stenetorp, P., S.~Pyysalo, G.~Topi{\'c}, T.~Ohta, S.~Ananiadou, y J.~Tsujii.
\newblock 2012.
\newblock brat: a web-based tool for {NLP}-assisted text annotation.
\newblock En F.~Segond, editor, {\em Proceedings of the Demonstrations at the 13th Conference of the {E}uropean Chapter of the Association for Computational Linguistics}, p\'aginas 102--107, Avignon, France, Abril. Association for Computational Linguistics.

\bibitem[\protect\citename{T{\"o}rnberg}2023]{tornberg2023use}
T{\"o}rnberg, P.
\newblock 2023.
\newblock How to use llms for text analysis.
\newblock {\em arXiv preprint arXiv:2307.13106}.

\bibitem[\protect\citename{Vaswani \bgroup et al.\egroup }2017]{NIPS2017_3f5ee243}
Vaswani, A., N.~Shazeer, N.~Parmar, J.~Uszkoreit, L.~Jones, A.~N. Gomez, L.~u. Kaiser, y I.~Polosukhin.
\newblock 2017.
\newblock Attention is all you need.
\newblock En I.~Guyon U.~V. Luxburg S.~Bengio H.~Wallach R.~Fergus S.~Vishwanathan, y R.~Garnett, editores, {\em Advances in Neural Information Processing Systems}, volumen~30. Curran Associates, Inc.

\bibitem[\protect\citename{Vila}2015]{vila2015ontoloxias}
Vila, T.~V.
\newblock 2015.
\newblock {\em Ontolox{\'\i}as e traduci{\'o}n biom{\'e}dica: creaci{\'o}n dunha base de co{\~n}ecemento terminol{\'o}xico sobre os erros innatos do metabolismo en franc{\'e}s e espa{\~n}ol}.
\newblock {Ph.D.} tesis, Universidade de Vigo.

\bibitem[\protect\citename{Villaplana, Mart{\'\i}nez, y Montalvo}2023]{villaplana2023improving}
Villaplana, A., R.~Mart{\'\i}nez, y S.~Montalvo.
\newblock 2023.
\newblock Improving medical entity recognition in spanish by means of biomedical language models.
\newblock {\em Electronics}, 12(23):4872.

\bibitem[\protect\citename{Wadud \bgroup et al.\egroup }2022]{wadud2022can}
Wadud, M. A.~H., M.~M. Kabir, M.~F. Mridha, M.~A. Ali, M.~A. Hamid, y M.~M. Monowar.
\newblock 2022.
\newblock How can we manage offensive text in social media-a text classification approach using lstm-boost.
\newblock {\em International Journal of Information Management Data Insights}, 2(2):100095.

\bibitem[\protect\citename{Xu \bgroup et al.\egroup }2022]{app122211709}
Xu, J., X.~Xi, J.~Chen, V.~S. Sheng, J.~Ma, y Z.~Cui.
\newblock 2022.
\newblock A survey of deep learning for electronic health records.
\newblock {\em Applied Sciences}, 12(22).

\end{thebibliography}

\appendix

\section{Experimental details}

In this section, we explain and describe in detail the results obtained and the procedure to reproduce the experiment.

\subsection{Algorithms for extracting information from ontologies}
In Algorithm \ref{alg1}, we explain the process of extracting information from the ontologies to add relationships between each report and the disease associated with it. 
\begin{enumerate}
    \item For each disease, we translate it into English using the Google Translate API.
    \item Then, for each ontology in our list we have different relationships, so it is important to use all of them. From the English translated disease name we have three options:
    \begin{enumerate}
    
        \item If SNOMED, we extract the site of the affected body and skin.
        \item If UMLS, we extract the by disease.
        \item If ICD10, we extract the severity of the disease by interpreting information about whether it is a major or minor condition, and whether it carries morbidity.
    \end{enumerate}
\end{enumerate}

\begin{algorithm}
	\caption{Relation extraction} 
    \label{alg1}
	\begin{algorithmic}[1]
        \State load \textit{pymedtermino} library
		\For {$diseases=1,2,\ldots,M$}
            \State translate {$diseases$} with Google API
			\For {$ontology=1,2,\ldots,N$}
                \If{$SNOMED$}
                    \State {$diseases.getType()$}
                \ElsIf{$UMLS$}
                    \State{$diseases.getLocation()$}
                \Else
				    \State $diseases.getAffection()$
                    \If{$has(minor)$}
                        \State{$SetTo(light)$}
                    \ElsIf{$has(major)$}
                        \State{$SetTo(important)$}
                    \ElsIf{$has(morbidity)$}
                        \State{$SetTo(deadly)$}
                    \Else
                        \State{$SetTo(inoffensive)$}
                    \EndIf
                \EndIf
			\EndFor
		\EndFor
	\end{algorithmic} 
\end{algorithm}

\subsection{Cascade models algorithm}

In this section, we explain the operation of cascade models (Algorithm \ref{alg2}): 
\begin{enumerate}
 \item We perform the ordered combination of the three relationships we have, with batches of size 1 or 3.
    \item For each of those packets, we load the dataset and tokenize it.
    \item For each element of each packet, we train a model in supervised mode to learn that element.
    \item Once each model is trained, its prediction of the item is concatenated with the original input, and serves as a new input to train the model to predict the next item in the packet.
    \item We train another model with the next element of the bundle and the new input.
    \item When all the elements have been learned by the different cascade models, a last model is trained with the original input enriched by the last output (with all the predicted elements) to predict the disease.
    \item Once all the packages are processed, the models are compared and the best one is chosen.
\end{enumerate}

\begin{algorithm}
	\caption{Cascade models}
 \label{alg2}
	\begin{algorithmic}[1]
        \State $combi \gets []$
		\For {$iter1=1,2,\ldots N$}
            \For{$iter2=1,2,\ldots M$}
		          \State $combi \gets iter1,iter2$
            \EndFor
        \EndFor
	    \For{$relations$ in $combi$}
            \State $input \gets string(medicalRecords)$
            \State $input.tokenize()$
            \For{$relation$ in $relations$}
                \State $output \gets model.train(relation)$
                \State {$input$ $\gets$ {$input + output$}}
            \EndFor
            \State $output \gets$ $Model.train(diseases)$
        \EndFor
        \State $computeAccuracy()$
        \State $getBestModel()$
	\end{algorithmic} 
\end{algorithm} 

\subsection{Training setup} \label{hyper_and_stuff}
The following model hyper-parameters have been identified as tunable:
\begin{itemize}
 \item Batch size. Number of training samples to be processed through the network in a single iteration before the model weights are updated.
 \item Learning rate. Controls how much the model weights are adjusted in response to the error computed at each iteration of training.
 \item Number of epochs. Number of times the learning algorithm will work through the entire training data set.
\end{itemize}

For each of these hyperparameters, different possible values have been considered and, after a \textit{grid search} process, it has been determined that the values that provide the best results are \textit{batch size} 64, \textit{learning rate} 0.001 and \textit{epochs} 10.

The experiments were conducted using an NVIDIA GeForce RTX 2080 Ti 12GB and an NVIDIA RTX A6000 50GB. We used Pytorch 2.2.1 library with CUDA 12.1. Each training took between 3 and 5 minutes given the limited data in the corpus. In total, counting the \textit{grid search}, the experiments took just over 96 hours to obtain the results presented in the paper. 

\subsection{Detailed results of the intermediate models} 
\label{res_inter}
In this section, we present the results of various combinations in the Table \ref{interm-fulltable} that were carried out to predict the pathology in the best possible way. Dealing with non-repeating variations, the total number of possible combinations is of the form: 
\[
\sum_{k=1}^{n} V_n^k = \sum_{k=1}^{n} \frac{n!}{(n-k)!}.
\]
For \( n = 3 \), the total number of combinations (variations) of sizes 1 to 3 without repetition is:

\[
\frac{3!}{(3-1)!} + \frac{3!}{(3-2)!} + \frac{3!}{(3-3)!} = 15.
\]

\begin{table*}[h]
\begin{center}
\begin{tabular}{|l|l|l|l|}
\hline
Information category & Accuracy &  Micro F1 & Macro F1 \\
\hline
\textbf{t} & 0.57 & 0.56  & 0.38        \\
\textbf{gr} & 0.57 & 0.56  & 0.41     \\
\textbf{sit} & 0.68 & 0.67  & 0.59     \\
\hline
\textbf{gr} $\rightarrow$ \textbf{t} $\rightarrow$ \textbf{sit}   & 0.62 & 0.61  & 0.51     \\\textbf{gr} $\rightarrow$ \textbf{sit} $\rightarrow$ \textbf{t}   & 0.66 & 0.66  & 0.58     \\
\textbf{t} $\rightarrow$ \textbf{gr} $\rightarrow$ \textbf{sit}  & 0.70 & 0.68  & 0.58   \\
\textbf{t} $\rightarrow$ \textbf{sit} $\rightarrow$ \textbf{gr}   & 0.70 & 0.68  & 0.57     \\\textbf{sit} $\rightarrow$ \textbf{t} $\rightarrow$ \textbf{gr}   & 0.69 & 0.67  & 0.58     \\
\textbf{sit} $\rightarrow$ \textbf{gr} $\rightarrow$ \textbf{t}  & \textbf{0.72} & \textbf{0.71 } &\textbf{0.62}     \\
\hline
\end{tabular}
\end{center}
\caption{\label{interm-fulltable} Table with intermediate results for type, severity and site of each disease.}
\end{table*}

\subsection{Comparison between the results of the different approaches}

A comparison of the complete OR method with the information of the ontologies (model A) and the method based on the bsc-bio-ehr-es \textit{vanilla} model without any added information (model B) has been carried out as a reference of the developments presented in this research, with special emphasis on the improvements provided by using the \textit{top-k} view, and seeing in which situations these changes have led to an improvement in the results. This comparison is presented in Table \ref{tabla_umbrales}.

For model A, the accuracy and F1-score metrics generated without the use of \textit{top-k} correspond to 0.82 and 0.73, which have been improved with the \textit{top-k} approach to 0.86 and 0.85 values. Similarly, it has been observed how model B goes from values 0.52 and 0.42 for these metrics to 0.67 and 0.61 with the inclusion of the \textit{top-k} approach. Although in model A there are evident improvements, it is in case B where there is a more substantial improvement.

\begin{table*}[h]
\begin{center}
\small

\begin{tabular}{|l|l|l|l|}
\hline
Pathology & F1 \textit{vanilla} & F1 PR & F1 OR \\
\hline
acne & 0.43 & 0.86 & 0.94 \\
basal cell carc. & 0.60 & 0.70 & 0.92  \\
psoriasis & 0.67 & 0.81 & 0.87 \\
melanocytic nevus & 0.52 & 0.72 & 0.93\\
actinic keratosis & 0.49 & 0.63 & 0.83\\
squamous cell carcinoma & 0.43 & 0.52 & 0.86\\
eczema & 0.45 & 0.59 & 0.62\\
rosacea & 0.00 & 0.37 & 0.55 \\
solar lentigo & 0.54 & 0.54  & 0.55\\
solar lentigo & 0.54 & 0.65 & 0.97\\
lichen scleroatrophicus & 0.73 & 0.69 & 0.82\\
fibroma & 0.57 & 0.50 & 0.87\\
sore & 0.64 & 0.69 & 0.78\\
melanoma & 0.43 & 0.66 & 0.86\\
alopecia areata & 0.74 & 0.80 & 0.80 \\
atopic dermatitis & 0.45 & 0.47 & 0.74 \\
squamous cell carcinoma & 0.40 & 0.60  & 0.91\\
seborrhoeic keratosis & 0.44 & 0.67 & 0.97\\
undiagnosed & 0.31 & 0.37 & 0.77  \\
juvenile acne & 0.04 & 0.00 & 0.63 \\
warts & 0.57 & 0.82 & 0.98\\
chronic urticaria & 0.22 & 0.71 & 0.87\\
haemangioma & 0.86 & 0.77 & 0.97\\
atypical melanocytic nevus & 0.00 & 0.00 & 0.91 \\
dermatofibroma & 0.53 & 0.53 & 0.95\\
ulcer & 0.00 & 0.55 & 0.70\\

\hline
\end{tabular}

\end{center}
\caption{\label{disease-table_full} Table with the metrics (without \textit{top-k}) for the 25 most frequent pathologies.}
\end{table*}

For model B, the main improvement occurs in cases where the disease is ``acquired melanocytic nevus'' and ``melanocytic nevus'' is predicted (11.89\%). Acquired and congenital melanocytic nevi share several features, including appearance, histology, genetics, and development. However, they differ in the timing of their appearance, with congenital nevi being present at birth or in the first few weeks of life, whereas acquired nevi appear throughout life. This illustrates how not only does the \textit{top-k} approach improve the accuracy of the models, but some of the most common errors are understandable given the meaning of the terms confused. 

It is also worth noting that the \textit{top-k} approach allows in several cases to improve the prediction that includes situations ``without diagnosis'', providing such an option in cases where the base approach does not contemplate it. It is also noteworthy how in both methods there is a high number of cases of confusion between diagnoses linked to carcinoma (``basal cell carcinoma'', ``squamous cell carcinoma'') and keratosis (``actinic keratosis'', ``seborrheic keratosis''). In this case they are not similar pathologies, so the inclusion of \textit{top-k} would provide the clinician with an alternative on which to assess which is the appropriate option with his or her expert knowledge. 

\subsection{Model results with different frequency thresholds for each pathology.} \label{an_umbral}
The Table \ref{tabla_umbrales} shows the results of both method A and B, and demonstrates that the threshold of 61 minimum examples per category is the optimal one to keep a maximum of categories without losing the efficiency of the classification models.

\begin{table*}[h]
\begin{center}
\resizebox{\textwidth}{!}{
\begin{tabular}{|l|l|l|l|l|l|l|l|}
\hline
Model & threshold & Num. classes & Acc. & Micro F1-sc. & Macro F1-sc & Acc. \textit{top-k} & F1-sc. \textit{top-k}\\
\hline

B & 2 & 173 & 0.39 & 0.34 & 0.08 & 0.54 & 0.13\\
B & 10 & 76 & 0.41 & 0.41 & 0.12 & 0.58 & 0.27\\
B & 25 & 44 & 0.46 & 0.43 & 0.24 & 0.59 & 0.46\\
B & 50 & 27 & 0.48 & 0.46 & 0.38 & 0.66 & 0.63 \\
\textbf{B (ours)} & 61 & 25 & 0.52 & 0.50 & 0.42 & 0.67 & 0.61\\
B & 75 & 20 & 0.51 & 0.49 & 0.44 & 0.69 & 0.71\\ 
B & 100 & 15 & 0.55 & 0.54 & 0.51 & 0.71 & 0.77 \\
\hline\hline
A & 2 & 173 & 0.68 & 0.62 & 0.14 & 0.80 & 0.28\\
A & 10 & 76 & 0.72 & 0.66 & 0.25 & 0.86 & 0.52\\
A & 25 & 44 & 0.77 & 0.73 & 0.48 & 0.90 & 0.81\\
A & 50 & 27 & 0.83 & 0.80 & 0.72 & 0.91 & 0.86\\
\textbf{A (ours)} & 61 & 25 & 0.84 & 0.82 & 0.75 & 0.92 & 0.90\\
A & 75 & 20 & \textbf{0.87} & 0.85 & 0.80 & 0.94 & 0.91\\
A & 100 & 15 & \textbf{0.87} & \textbf{0.87} & \textbf{0.85} & \textbf{0.96} & \textbf{0.92}\\

\hline
\end{tabular}}
\end{center}
\caption{\label{tabla_umbrales} Metrics obtained with different thresholds of examples per pathology A is the method with the information from the ontologies and B the method with the bsc-bio-ehr-es \textit{vanilla}.}
\end{table*}

\section{Additional information} \label{an_ic}

\begin{figure*}[h]
 \centering
 \includegraphics[width=\linewidth]{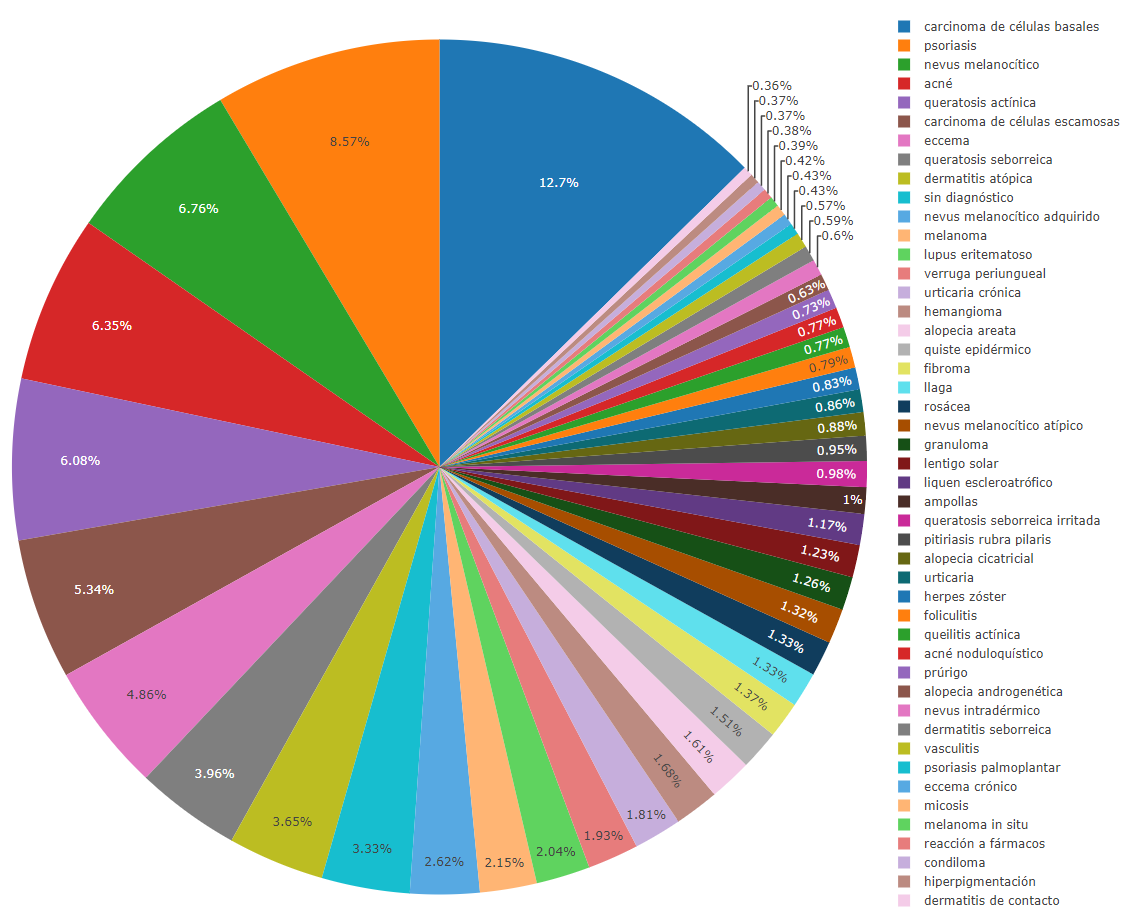}
 \caption{Distribution of diseases in the generated dataset}
 \label{distr_enf}
\end{figure*}

\begin{figure*}[h]
 \centering
 \includegraphics[width=\linewidth]{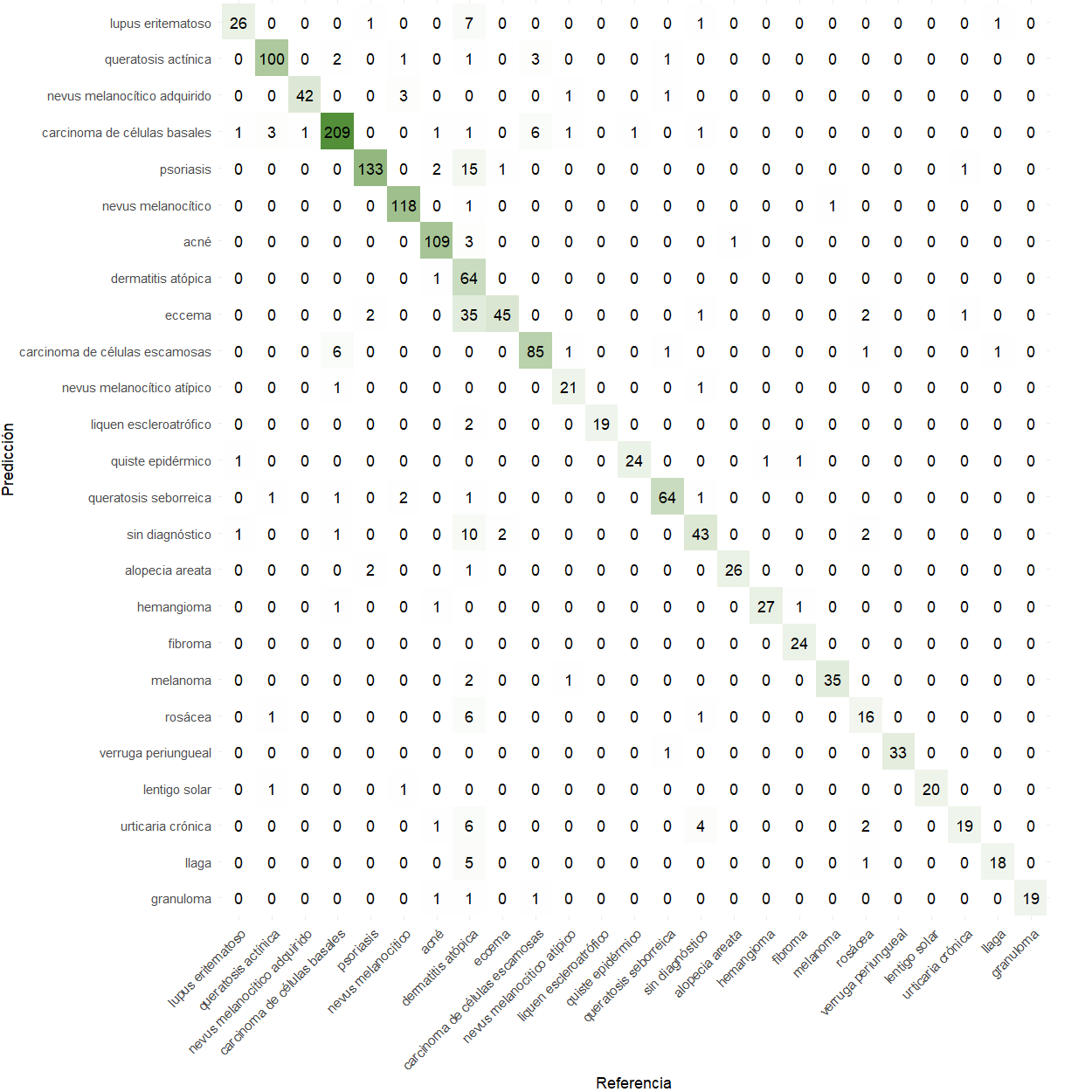}
 \caption{Confusion Matrix for model A}
 \label{cm_full}
\end{figure*}

\begin{figure*}[h]
 \centering
 \includegraphics[width=\linewidth]{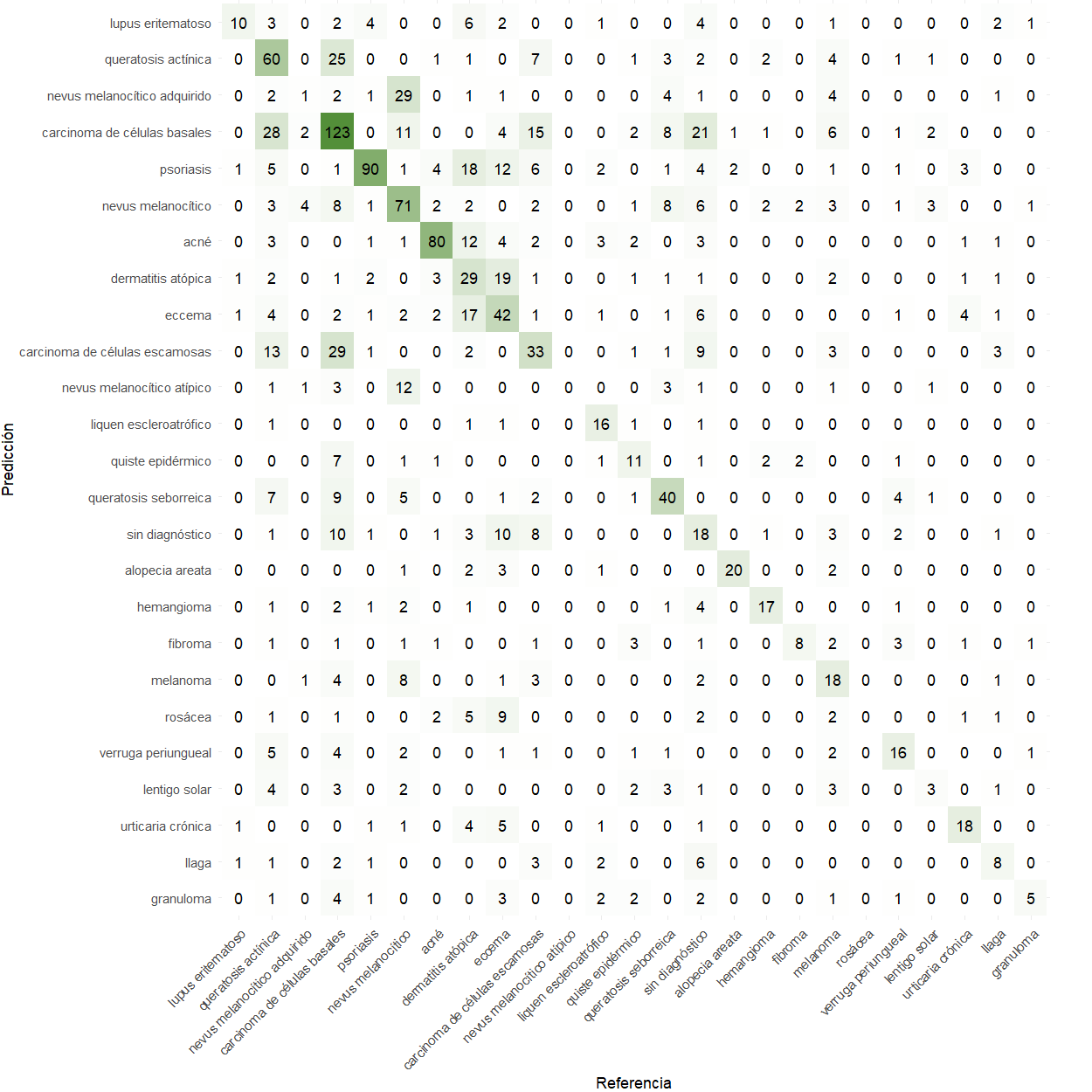}
 \caption{Confusion Matrix for model B}
 \label{cm_vanilla}
\end{figure*}

This section presents complementary information to support the understanding of the developments of this work. In particular, the following information is presented:

\begin{itemize}
    \item Table \ref{type_transl}. Extended version of Table \ref{type_transl}, with the complete list of diseases.
    \item Figure \ref{distr_enf}. Graphical representation of the distribution of diseases in the data set.
    \item Figures \ref{cm_full} and \ref{cm_vanilla}. Confusion matrices of methods A and B presented in Annex \ref{an_ic}, respectively. 
\end{itemize}

\begin{table*}[h]
\begin{center}
\begin{tabular}{|l|l|l|l|l|}
\hline
Disease & Type & Severity & Site & Frequency\\
\hline
basal cell carcinoma & neoplastic process & major & skin & 1124 \\
psoriasis & autoimmune process & harmless & extremities & 761 \\
melanocytic nevus & precancer & harmless & all & 600 \\
acne & disease & mild & all & 564 \\
actinic keratosis & precancer & harmless & skin & 540 \\
squamous cell carcinoma & neoplastic process & extreme & skin & 474 \\
eczema & disease & harmless & hand & 432 \\
seborrheic keratosis & benign tumor & harmless & skin & 352 \\
atopic dermatitis & disease & harmless & joints & 324 \\
undiagnosed & no disease & harmless & all & 296 \\
acquired melanocytic nevus & neoplastic process & harmless & extremities & 233 \\
melanoma & neoplastic process & extreme & all & 191 \\
lupus erythematosus & autoimmune process & extreme & connective tissue & 181 \\
periungual wart & infection & harmless & hand & 171 \\
chronic urticaria & symptom & harmless & all & 161 \\
hemangioma & benign tumor & mild & all & 149 \\
alopecia areata & autoimmune process & harmless & head & 143 \\
epidermal cyst & abnormality & mild & face & 134 \\
fibroma & benign tumor & mild & leg & 122 \\
sore & symptom & harmless & mouth & 118 \\
rosacea & disease & harmless & face & 118 \\
atypical melanocytic nevus & neoplastic process & important & torso & 117 \\
granuloma & infection & extreme & genitals & 112 \\
cutaneous lentigo & syndrome & mild & all & 109 \\
lichen sclerosus & autoimmune process & mild & genitals & 104 \\
blisters & symptom & harmless & hand & 89 \\
irritated seborrheic keratosis & disease & harmless & all & 87 \\
pityriasis rubra pilaris & autoimmune process & mild & joints & 84 \\
scarring alopecia & disease & harmless & head & 78 \\
urticaria & pathological function & mild & all & 76 \\
herpes zoster & infection & important & torso & 74 \\
folliculitis & disease & harmless & head & 70 \\
actinic cheilitis & precancer & mild & mouth & 68 \\
nodulocystic acne & infection & mild & face & 68 \\
prurigo & symptom & harmless & head & 65 \\
androgenetic alopecia & disease & harmless & head & 56 \\
intradermal nevus & precancer & harmless & skin & 53 \\
seborrheic dermatitis & autoimmune process & harmless & face & 52 \\
vasculitis & autoimmune process & extreme & joints & 51 \\
palmoplantar psoriasis & disease & mild & extremities & 38 \\
chronic eczema & disease & harmless & hand & 38 \\
mycosis & infection & important & all & 37 \\
melanoma in situ & neoplastic process & harmless & all & 35 \\
drug reaction & poisoning & harmless & all & 34 \\
condyloma & infection & mild & genitals & 33 \\
hyperpigmentation & abnormality & harmless & all & 33 \\
contact dermatitis & disease & harmless & hand & 32 \\
\hline
\end{tabular}
\end{center}
\caption{\label{type_transl_an} Nomenclature generated from the characteristics extracted with \textit{Pymedtermino} with the most frequent diseases and the number of occurrences}
\end{table*}

\end{document}